\documentclass{article}

\usepackage{arxiv}

\usepackage[utf8]{inputenc} 
\usepackage[T1]{fontenc}    
\usepackage{hyperref}       
\usepackage{url}            
\usepackage{booktabs}       
\usepackage{amsfonts}       
\usepackage{nicefrac}       
\usepackage{microtype}      
\usepackage{lipsum}

\title{Object Detection with Deep Reinforcement Learning}

\author{
  Manoosh Samiei\\
  Electrical and Computer Eng. Department\\
  McGill University\\
  \texttt{manoosh.samiei@mail.mcgill.ca} \\
   \And
  Ruofeng Li \\
  Electrical and Computer Eng. Department\\
  McGill University\\
  \texttt{ruofeng.li@mail.mcgill.ca} \\
}

\begin{document}
\maketitle

\begin{abstract}
Object localization has been a crucial task in computer vision field. Methods of localizing objects in an image have been proposed based on the features of the attended pixels. Recently researchers have proposed methods to formulate object localization as a dynamic decision process, which can be solved by a reinforcement learning approach. In this project, we implement a novel active object localization algorithm based on deep reinforcement learning. We compare two different action settings for this MDP: a hierarchical method and a dynamic method. We further perform some ablation studies on the performance of the models by investigating different hyperparameters and various architecture changes. 
\end{abstract}


\section{Introduction}

Object classification and object detection are two of the most important tasks in computer vision. An object classification algorithm identifies which objects are present in an image. While, an object detection (localization) algorithm not only indicates which objects are present in the image, but also outputs bounding boxes to indicate the location of the objects inside the image. Object detection has significant applications in real life. In self-driving vehicles, object detection is used to detect cars, pedestrians, traffic lights, road signs, etc. around the vehicle to help it decide its next actions. Object tracking is another application that is vastly used in traffic monitoring, activity recognition, ball tracking in sports, surveillance and security i.e. tracking a person in a video. Face and iris detection are used as powerful means of identity verification. Also, detection of organs is crucial in medical imaging to assist the clinicians in diagnosis, therapy planning and image-guided interventions, by detecting and analyzing the organs of a patient. 

In this project we train a model to localize objects in images using a deep reinforcement learning approach. We formulate object detection problem as a Markov-Decision Process(MDP), and try to find the salient parts of an image that are more probable to contain a target object, and then zoom on them. This process is repeated until a tight bounding box is found around the target object.

We implement two different action settings for this MDP. In the first setting, a hierarchical method proposed by \cite{Bellver2016HierarchicalOD}, the agent chooses to focus on one of the 5 sub-regions of the image (ie. top-left, top-right, bottom-left, bottom-right, center) at each time step. In the second setting, a dynamic method proposed by \cite{iccv}, the agent deforms a bounding box using simple transformation actions (horizontal moves, vertical moves, scale changes, and aspect ratio changes) at each step to find the specific location of an object in the image. These methods are class-specific and can find a specific class of object at a time. Both methods are able to localize objects within 10 steps (region proposals). As suggested by \cite{iccv}, attending smaller number of regions proposals is a superiority of deep reinforcement learning methods over R-CNN method which is a baseline model in object detection. We also attempt to enhance the hierarchial method, by changing the network's architecture, multiple setups, and hyperparameters. To train and evaluate the model, we use Pascal VOC2012 dataset. 

\subsection{Background} 

The first step in object detection task is to extract features from the input image, using convolutional neural networks or CNN. Convolutional layers reduce the dimensions of input data by performing a convolution operation on them, which makes them appropriate for handling image data. We use a pre-trained VGG16 network to extract the features from the input images as suggested by \cite{Bellver2016HierarchicalOD}. We use \textbf{deep q networks}, proposed by \cite{mnih2013atari}, instead of basic q-learning updates to estimate state-action values (q), since DQN fits better to large state spaces where simple q-learning algorithm fails to perform efficiently. DQN is just an artificial neural network which estimates the optimal q function. DQN accepts state from a given environment as input and for each given state, the network outputs estimated q values for each action that can be taken from that sate. The objective of this network is to approximate the optimal q function that is derived from bellman equation \ref{eq:eq1}.

\begin{equation}
q_*(s,a) = E[R_{t+1}+\gamma \max_{a'}{q_{*}(s',a')}]
\label{eq:eq1}
\end{equation}

The loss from the network is calculated by comparing the outputted Q values to the true optimal Q values. The objective of network is to minimize this loss. After the loss is calculated, the weights of the network are updates via stochastic gradient descent and back propagation. This process is done repeatedly for all states of the environment until the losses are sufficiently minimized and an approximate optimal q function is obtained. This process process is similar to one-step Q-learning update rule in \ref{eq:eq2}. 

\begin{equation}
w_{t+1} = w_{t} + \alpha[R_{t+1} + \gamma\max_{a}{q(s_{t+1},a,w_t)} - q(s_t,a_t,w_t)]\nabla q(s_t,a_t,w_t)
\label{eq:eq2}
\end{equation}

Training a DQN needs a special setting called experience replay. With experience replay we store agent's experiences, defined as: $e_t = (current\text{ }state, action, reward, next\text{ }state)$, at each time step in replay memory. If the network only learns from consecutive samples of experiences as occurred sequentially in the environment, the samples would be highly correlated and lead to inefficient learning. Random samples from replay memory are then used to train DQN in order to break the correlation between consecutive samples. Using the estimated q values from deep q network, we can find an optimal policy for object region proposals to localize a specific category of objects in that image. 

\section{Method}

\subsection{Markov decision process formulation}

We define object detection problem as a Markov decision process, in which our goal-oriented agent interacts with a visual environment that is our image. At each time step in training phase, the agent applies a transformation to a bounding box in image, a positive or negative reward is assigned to agent based on the new bounding box, q values are updated based on the the state, action and reward, and DQN is trained. During testing, the agent does not receive rewards and does not update the model either, it just follows the learned policy. We implemented two different formulations from two papers: a dynamic method proposed by \cite{iccv} and a hierarchical method \cite{Bellver2016HierarchicalOD}. The MDP formulation of these two methods differ in their action setting and to some extent their state representation. We will describe the components of these MDPs as follows.

\subsubsection{Actions for Hierarchical Method}
\begin{figure}[!htb]
  \centering
  \includegraphics[width=0.4 \textwidth]{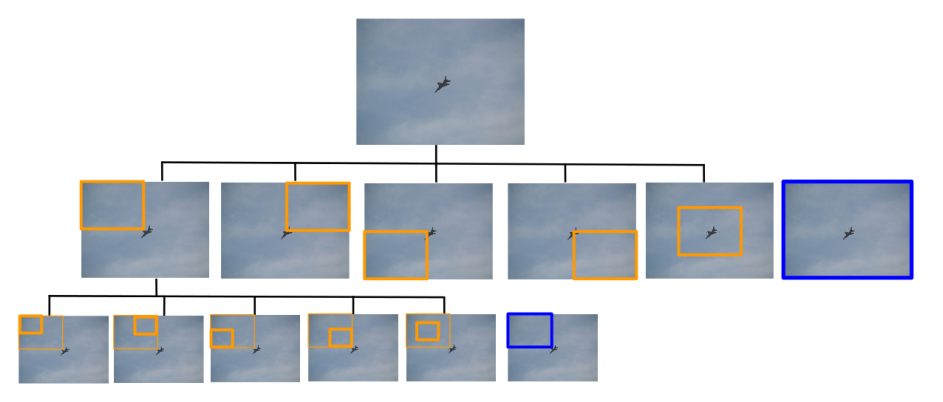}
  \caption{ Illustration of hierarchical actions in MDP. Image credit \cite{Bellver2016HierarchicalOD}}
  \label{fig:action_h}
\end{figure}

Generally in both hierarchical and dynamic method, there are several \textbf{movement} actions and a \textbf{terminal} action. In hierarchical method, there are five movement actions which correspond to five sub-regions in the current bounding box: the four regions representing the four quadrants plus a central region. There is also a terminal action called trigger, which should be chosen when the agent reaches its goal of finding a right bounding box around the target object. Figure \ref{fig:action_h} can further clarify why this method is called hierarchical. 

\subsubsection{Actions for Dynamic Method}
In dynamic method, there are eight transformation actions that change the geometry and location of the bounding boxes, and one trigger action. The eight transformations can be divided into four categories, which modify the horizontal level, vertical level, scale and aspect ratio of the bounding box. Figure \ref{fig:action} can further clarify these transformations. Any transformation makes a discrete change to the box by a relative factor ($\alpha$) to its current size, using equations \ref{eq:act}.

\begin{equation}
\begin{aligned}
    \alpha_{w} = \alpha (x_2 - x_1)
    && \alpha_h = \alpha (y_2 - y_1)
\label{eq:act}
\end{aligned}
\end{equation}

\begin{figure}[!htb]
  \centering
  \includegraphics[width=0.4 \textwidth]{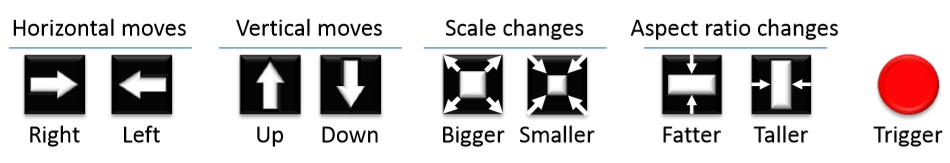}
  \caption{ Illustration of dynamic actions in MDP. Image credit \cite{iccv}}
  \label{fig:action}
\end{figure}

Where $\alpha \in [0,1]$ and $x_1$, $y_1$, $x_2$, and $y_2$ are the coordinates of the top-left corner and the right-bottom corner of the current placed bounding box, respectively. Then according to the type of transformation action taken, $\alpha_{w}$ and $\alpha_{h}$ are added to or subtracted from the current bounding box coordinates. 
 
\subsubsection{States}
The states of this MDP, are constructed of a feature vector which summarizes the image information of the current found bounding box, and a history vector of several previous taken actions. In both hierarchical and dynamic method, the feature vector is extracted by feeding the current bounded region of the image into a pre-trained VGG-16 network. Since the VGG-16 network requires for inputs with the same dimension, all the attended regions are  resized to 224x224 dimension. In hierarchical method, the output of the last pooling layer which is 25088-dimensional is used as feature vector; while in dynamic method, the output of the first dense layer which is 4096-dimensional is used as feature vector. 
In hierarchical method, the action history vector \emph{h} contains 4 past actions of agent in a one-hot encoded format, and as we have 6 actions, the binary one-hot encoded history vector has 24 elements. In dynamic method, the history vector \emph{h} contains 10 past actions of agent, and as we have 9 actions, the history vector has 90 elements. By adding this history vector as part of the state representation, the agent is expected to avoid repetitive actions in its search paths. 

\subsubsection{Intersection-over-Union (IoU)}
Before explaining the reward formulation of our MDP, we need to define a metric called Intersection-over-Union (IoU). We mainly exploit the Intersection over Union (IoU) between the predicted bounding box and the ground truth mask to evaluate the performance of our model. The IoU is defined as the ratio of the overlapped area between the predicted bounding box and the ground-truth bounding box over the area encompassed by both of them. More specifically, given the predicted bounding box $b$ and the ground truth bounding box $g$, the IoU between $b$ and $g$ is expressed as the following formula:

\begin{equation}
\begin{aligned}
IoU(b,g)= \frac{area(b\cap g)}{area(b\cup g)}
\label{eq:iouu}
\end{aligned}
\end{equation}

\subsubsection{Reward function R}
The reward function needs to be defined corresponding to the actions taken. In the proposed formulation, the reward function $R$ is given based on the improvement of the Intersection-over-Union (IoU) between the predicted and the ground truth bounding box, during the transition between states. Thus, for non-terminal actions, the reward is set to 1 if the state transition improves the IoU and -1 otherwise, as suggested by below formula:

\begin{equation}
\begin{aligned}
R_a(s,s')=sign(IoU(b',g) - IoU(b,g))
\label{eq:Reward}
\end{aligned}
\end{equation}

For the terminal action (trigger), the reward scheme is different since the terminal action does not change the geometry of the bounding box. As a result, the reward is defined based on whether the current IoU exceeds the pre-set threshold ($\tau$) or not. The terminal reward is set to 3 if the current IoU is larger than the threshold and -3 otherwise, as suggested by below formula:

\begin{equation}
\begin{aligned}
R_w(s,s')= 
    \begin{cases} 
          +\eta \text{ if }IoU(b,g)\geq \tau \\
          -\eta \text{ otherwise}
       \end{cases}
\label{eq:terminal}
\end{aligned}
\end{equation}

Where $\eta$ was chosen to be 3 for both dynamic and hierarchical methods and $\tau$ (trigger threshold) was chosen to be 0.6 in dynamic method \cite{iccv} and 0.5 for hierarchical method \cite{Bellver2016HierarchicalOD}.

\subsection{Model Architecture}
 
\begin{figure}[!htb]
  \centering
  \frame{\includegraphics[width=0.337 \textwidth]{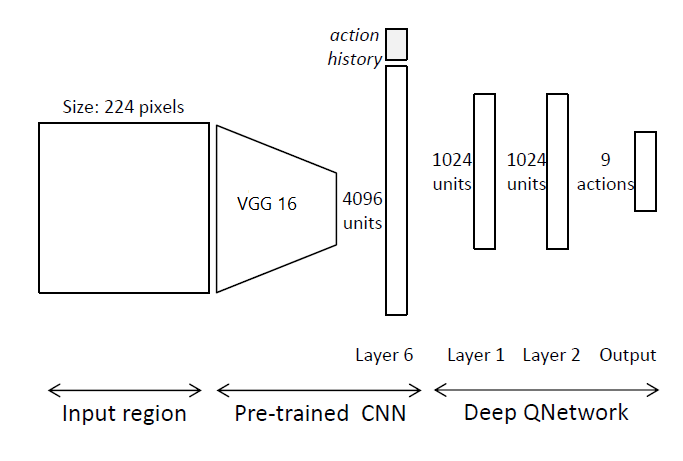}}
  \frame{\includegraphics[width=0.46 \textwidth]{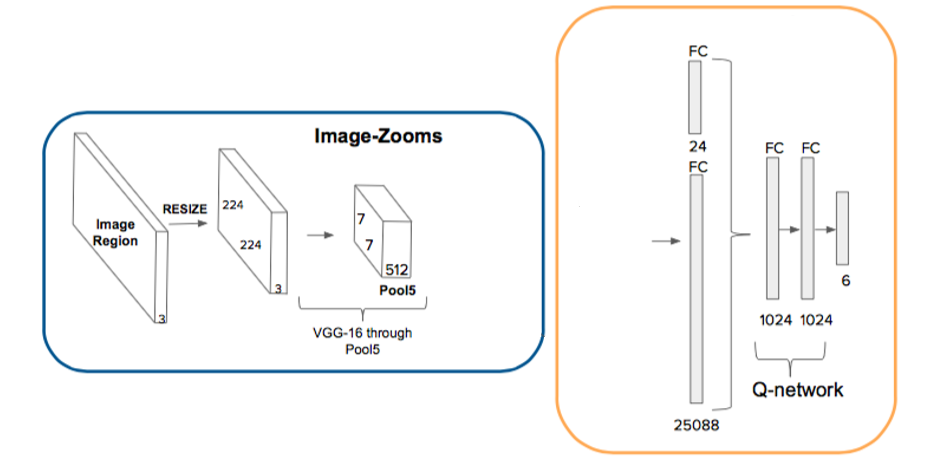}}
  \caption{Architecture of dynamic model in the left \cite{iccv} and hierarchical model in the right \cite{Bellver2016HierarchicalOD}}
  \label{fig:fig2}
\end{figure}

The architecture of both hierarchical and dynamic models that we implemented can be seen in \ref{fig:fig2}. In the original paper of dynamic method \cite{iccv}, the authors use a 5-layer pre-trained convolutional neural network instead of a VGG16 network to extract features; however, for better comparability of the two approaches we used a pretrained VGG16 for both models. VGG16 network is composed of 13 convolutional layers, 5 max pooling layers, and 3 dense layers. In hierarchical method, the output of the last max pooling layer is concatenated with 24-dimensional action history vector which forms our state representation. In dynamic method, the output of first dense layer in vgg16 is concatenated with a 90-dimensional action history vector and forms our state representation. This final array is then fed to a deep q network which is composed of 3 dense layers. The first two layers are composed of 1024 units and the last dense layer contains q-values for each of our actions, 6 units in hierarchical method and 9 units in dynamic method. It is worth mentioning that in DQN structure the output layer is not followed by any non-linear activation function since we need the raw non-transformed q values from the network. In our implementation, we used a linear activation function for the output layer.

\subsection{Dataset}\label{AA}

The dataset applied in this project is PASCAL Visual Object Classes (VOC) 2012 dataset. This dataset is composed of images for twenty object categories, such as person, bird, cat, cow, dog, etc. There are around 12,000 images in this dataset, for which the ground truth bounding boxes for model training and testing are provided.

In this dataset, there are several text files for each object category, which provide the name of all images along with a number (1 or -1) to show whether each image contains an object from that category. This helps the users to train the model with specific classes. Since the model (in both dynamic and hierarchical method) is class-dependent, we trained the model with images containing objects from one class at each time.

\section{Results and Discussions}

\subsection{Hierarchical Method}

In \cite{Bellver2016HierarchicalOD} authors test another variation of hierarchical model, called pool45-Crops,that only extracts features for the whole image, and then reuses some regions of feature map instead of extracting new features for each predicted bounding box. However, they discuss in their paper that this change degrades the performance. Therefore we did not implemented this approach and we only focused on improving the better model called image-zooms by performing hyper-parameter search, and changing model structure and set-ups. Due to the limitation of the time, we conducted hyperparameter search experiments only on aeroplane class. We also evaluated the performance of our best model on seven other classes to examine the generalization of our method. 

Also in all experiments, the initial value for exploration parameter, $\epsilon$, is set to 0.9 (in dynamic method it is set to 1), and at the end of each epoch it is decreased by 0.1, until its value reaches 0.1 when it stays constant for the rest the epochs.

To evaluate the performance of different settings we use average IoU (Intersection over Union) over all testing images. For images with multiple objects from the target class, we measured the IoU between the prediction and all ground truth boxes and picked the maximum IoU. 

In all experiments the agent is at least trained for 10 epochs, and the models of all epochs are stored separately and tested. The highest average IoU over these epochs is used for comparison.

The highest average IoU we obtain using the default settings, proposed by \cite{Bellver2016HierarchicalOD}, is \textbf{0.3997}. In the following we investigate the effect of several changes in the original setting of the model.

\subsubsection{Double Q Learning}

As suggested by \cite{HasseltGS15}, one problem associated with deep q networks, or generally q learning algorithms, is the over estimation of Q values. In other words, as our Q-values move closer to their targets, the targets continue to move further because we are using the same network to calculate both of these values. To solve this issue, we implemented two separate network, one to estimate the target value at each step called target network and one to learn policy (q values) called policy network. The target network is a clone of the policy network. Its weights are frozen with the original policy network’s weights, and we update the weights in the target network to the policy network’s new weights every certain amount of time steps. We tested updating the weights of target network to policy network with three different number of time steps, i.e. every 5 steps, every 10 steps, and every 1 step. As it can be seen in table \ref{double}, updating weights of target network with bigger time intervals, leads to better results.  

\begin{table}[htbp]
\caption{Number of steps between updating target network weights}
\begin{center}
\begin{tabular}{|c|c|}
\hline
\textbf{Updating weights interval} & \textbf{Highest Average IoU}\\
\hline 
every 10 step & 0.4151 \\
\hline
every 5 step  &  0.3899\\
\hline
every step  &  0.3882\\
\hline
\end{tabular}
\label{double}
\end{center}
\end{table}

\subsubsection{Overlapping vs Non-overlapping Sub-regions}

In the original setting of model, each sub-region is 3/4 of its ancestor bounding box, in both width and height. This setting leads all 5 subregions to have overlap with each other. We investigated a non-overalpping setting, in which each subregion's width and height is 1/2 of its ancestor box, and therefore there is no overlap between the subregions. Image \ref{fig:action_h} is showing a non-overlapping case of hierarchical method. To make this change in the code, we set a parameter called 'scale subregion' to 1/2. There is another parameter in the code, called 'scale mask', that controls the movements of sub-regions to 5 different locations in image, i.e. top-left, top-right, bottom-left, bottom-right, and center. The amount of movement of subregion, is determined by the multiplication of scale mask and scale subregion parameters. Therefore, in non-overlapping case, 'scale mask' should be equal to 1, so that by multiplying it with 1/2, we move each mask by 1/2 and cover the whole area of ancestor box. We tested three other modifications to these two parameters, that can be seen in table \ref{scale}. When scale subregion is 1/2 and scale mask is 1/2 the masks move by 1/4 in ancestor box and therefore do not cover the whole surface of the box and are concentrated in the middle of the box. We also tested bigger subregions, each 4/5 of the previous box, both with full coverage of box and concentrated in the middle of box. Two results were obtained: first smaller windows (1/2) have better performance, second when windows are concentrated in the middle of the image they might have better performance due to most objects being presents relatively in the middle. In \cite{Bellver2016HierarchicalOD} authors suggest that overlapping windows perform better than non-overlapping case; while in our experiments we obtained the best result with the non-overlapping case.

\begin{table}[htbp]
\caption{Changing the size and movement of sub-regions in ancestor box}
\begin{center}
\begin{tabular}{|c|c|c|c|}
\hline
\textbf{Setting} & \textbf{Scale subregion} &  \textbf{Scale mask}& \textbf{Highest Average IoU}\\
\hline 
Non-overlapping & $1/2$  &  $1$ & \textbf{0.4599} \\
\hline
Smaller, Concentrated & $1/2$  &  $1/2$ & 0.4527 \\
\hline
Default & $3/4$ & $1/3$ & 0.3997 \\
\hline
Bigger, Concentrated & $4/5$  &  $5/16$ & 0.3629 \\
\hline
Bigger & $4/5$  &  $1/4$ & 0.3438 \\
\hline
\end{tabular}
\label{scale}
\end{center}
\end{table}

\subsubsection{Reward Setting}

To investigate the effect of different reward settings, we modified the original reward set up in two ways. One problem that we observed was that for some images the agent fails to choose trigger action on time; it continues making window smaller up to the maximum number of steps allowed in testing. Thus we decided to encourage agent to choose trigger action earlier, by increasing the value of reward assigned for the trigger action ($\eta$ in formula \ref{eq:terminal}). The results can be seen in table \ref{rewards}.

\begin{table}[htbp]
\caption{Changing the value of trigger reward}
\begin{center}
\begin{tabular}{|c|c|}
\hline
\textbf{Trigger action reward ($\eta$)} & \textbf{Highest average IoU}\\
\hline 
3 (default) & 0.3997 \\
\hline
6 &  0.2081\\
\hline
10 &  0.2252\\
\hline
\end{tabular}
\label{rewards}
\end{center}
\end{table}

In the second setting, we used another definition called recall instead of IoU for reward measurement. The formula for the recall is:
\begin{equation}
\begin{aligned}
Recall(b,g)= \frac{area(b\cap g)}{area(g)}
\label{eq:iouu}
\end{aligned}
\end{equation}

where, compared to IoU formula, the denominator is changed from the union of the predicted bounding box (b) and the ground truth (g) to just the area of the ground truth box (g). 
We experimented with three different values for recall threshold of choosing trigger action ($\tau$). Although we are training agent to maximize its recall, we measured its average IoU to be able to compare its performance with other settings. The highest average IoU up to 10 epochs corresponding to each experiment is shown in table \ref{recall}.

\begin{table}[htbp]
\caption{Changing the threshold of trigger recall}
\begin{center}
\begin{tabular}{|c|c|}
\hline
\textbf{Recall threshold} & \textbf{Average IoU}\\
\hline 
0.5 & 0.3916 \\
\hline
0.6 &  0.3793\\
\hline
0.7 &  0.3786\\
\hline
\end{tabular}
\label{recall}
\end{center}
\end{table}

From the results, we observe that the default setting with IoU as the evaluation method and trigger reward ($\eta$) equal to 3 performs the best. With higher trigger reward, the agent tends to trigger at the very early steps, thus resulting in a relatively low IoU. By changing the reward evaluation method from IoU to recall, the agent tends to search for actions that maximize the recall, i.e. it tends to predict larger windows instead of an accurate one around the object, giving a lower IoU in the end. 

\subsection{Fixing the target object during training}

In VOC2012 dataset, some images contain more than one object of the same class. In the original code published by the authors \cite{Bellver2016HierarchicalOD}, for multiple-target images, the agent changes the target during training based on the location of the current predicted box. In other words, the agent tries to find the object, with which it currently has higher IoU, and if it is searching for another object further away from its current bounding box in image, it will change its target object to a closer one. We tested the agent performance while fixing its target object, to investigate the effect of this behavior in agent's training. As a result of this change the agent's highest average IoU was increased to \textbf{0.4217}, an increase of around 0.02 compared to its original setting. We believe that changing target object can cause confusion for agent during training, and consequently fixing the target object improves its performance. 

\subsubsection {Training and test steps}

We tested two other values: 7 and 15 for the number of steps per epoch during training and testing time to investigate its effect on performance. We realized that the default setting with 10 number of steps per epoch works best during training. However during testing, decreasing the number of steps, leads to higher performance for some experiments. We argue that limiting the number of steps during test prevents the agent from making the predicted bounding box smaller, keeping a higher IoU (intersection) with the ground truth box. The results for the number of training steps can be seen in table \ref{steps}.

\begin{table}[htbp]
\caption{Changing the number of training steps per epoch}
\begin{center}
\begin{tabular}{|c|c|}
\hline
\textbf{Training Steps per epoch} & \textbf{Average IoU}\\
\hline 
10 (default) & 0.3997 \\
\hline
15 &  0.3340\\
\hline
7 &  0.3497\\
\hline
\end{tabular}
\label{steps}
\end{center}
\end{table}

\subsubsection{Discount rate $\gamma$}

We tested three different values for the discount rate ($\gamma$). The results of these experiments are shown in table \ref{gamma}.

\begin{table}[htbp]
\caption{Changing the value of discount rate $\gamma$}
\begin{center}
\begin{tabular}{|c|c|}
\hline
\textbf{Value of $\gamma$} & \textbf{Average IoU}\\
\hline 
0.1 & 0.2785 \\
\hline
0.5 &  0.3044\\
\hline
0.9 (default )&  0.3997\\
\hline
\end{tabular}
\label{gamma}
\end{center}
\end{table}

From the results we observe that with higher $\gamma$, the average IoU is higher. With a higher $\gamma$, the agent is more far-sighted, meaning that it gives great importance to the future rewards and acts in a way to maximize its long-term reward. Therefore, it chooses regions that in the end lead to higher intersection with the target object in image. While with a smaller $\gamma$, the agent tries to maximize its short-term reward and continues searching for the object, which causes the predicted window to become smaller and smaller, leading to a lower IoU in the end.

\subsubsection{Batch size}

During training, at each time step, we extract a fixed number of random experiences, called batch size, from replay memory to train our deep q network. We tested three different values for batch size, to investigate its effect on training. The results are shown in table \ref{batch}.

\begin{table}[htbp]
\caption{Changing the number of samples from replay memory}
\begin{center}
\begin{tabular}{|c|c|}
\hline
\textbf{batch size} & \textbf{ Average IoU}\\
\hline 
50 & 0.4041 \\
\hline
100 (default) &  0.3997\\
\hline
200 &  0.3170\\
\hline
\end{tabular}
\label{batch}
\end{center}
\end{table}

From the results, we observe that with higher random training samples, the performance degrades. This sounds surprising as we expected that with larger training samples at each step, DQN is trained better. One probable reason might be the higher probability of extracting correlated samples from replay memory. In these experiments, the size of replay memory was 1000 samples, and when extracting higher number of samples from memory (200 vs. 50), it is more probable to extract correlated samples, which as discussed earlier leads to inefficient training. 




\subsubsection{Trigger threshold}

We modified the IoU threshold ($\tau$) for choosing the trigger action illustrated in formula \ref{eq:terminal}. The results are shown in table \ref{tauuu}.

\begin{table}[htbp]
\caption{Changing the value of trigger threshold}
\begin{center}
\begin{tabular}{|c|c|}
\hline
\textbf{Trigger threshold} & \textbf{ Average IoU}\\
\hline 
0.5 (default) & 0.3997 \\
\hline
0.6 & 0.3585 \\
\hline
0.7 & 0.3835 \\
\hline
\end{tabular}
\label{tauuu}
\end{center}
\end{table}

From the results we observe that increasing the threshold of trigger action decreases the performance of the model. This could be due to that it is hard for the agent to find IoU scores of 0.6 and above, during training, and thus many times choosing trigger action leads to negative rewards, according to formula \ref{eq:terminal}. Consequently, choosing trigger action is discouraged in training and in the test phase, the agent fails to choose the trigger action at the right time, causing overly compressed predicted windows. 

\subsubsection{Size of action history vector}

The action history vector, as introduced in the Method section, is used to prevent the agent from repeating its previous actions during the search. We tested multiple values for the size of action history vector, in the setting of not changing the target object. In particular, we considered the state representation without action history, with one action history, and four action history. The results for these experiments are shown in table \ref{actionhist}. surprisingly, the case without any action history works the best. We argue that the action history vector in the state representation may not be well-understood by the agent. Action history vector contains only binary values and is concatenated to the output of vgg16, which contains numbers in a great range. Probably, the agent cannot interpret the action history part of the state representation. 

\begin{table}[htbp]
\caption{Changing the size of action history vector}
\begin{center}
\begin{tabular}{|c|c|}
\hline
\textbf{Size of action history vector} & \textbf{ Average IoU}\\
\hline 
without history &  0.4385 \\
\hline
1 action history &  0.4245\\
\hline
4 action history (default) &  0.4217\\
\hline
6 action history &  0.4053\\
\hline
\end{tabular}
\label{actionhist}
\end{center}
\end{table}

\subsubsection{Adding layer to DQN}

We also added one extra dense layer to our deep q network (i.e. 4 layer DQN) in double learning mode. The model performance was degraded with its best average IoU being \textbf{0.3524}. 

\subsubsection{Best-performance hierarchical model}

Among all experiments we performed, the non-overlapping windows had the highest average IoU (0.4599). However, when we looked at the images of different settings, we observed more reasonable predicted bounding boxes with the \textbf{fixed target object} setting. Sometimes, choosing trigger action in early steps maximizes the IoU while does not predict the location of object accurately. Due to time limitation, we could not test the combination of both of these methods, i.e. non-overlap + fixed target object, and we suggest this to be considered as a future work.

\subsubsection{Other object categories}

We tested the performance of the model, with the setting of fixing the target object, on 7 other object classes. The results can be seen in table \ref{class}. 

\begin{table}[htbp]
\caption{Different Object Categories}
\begin{center}
\begin{tabular}{|c|c|}
\hline
\textbf{Object class} & \textbf{ Average IoU}\\
\hline 
aeroplane &  0.4217 \\
\hline
bicycle &  0.4225\\
\hline
car &  0.2762\\
\hline
cat &  0.5172\\
\hline
bird &  0.2628\\
\hline
bus &  0.4437\\
\hline
dog &  0.4518\\
\hline
horse &  0.4149\\
\hline
\end{tabular}
\label{class}
\end{center}
\end{table}

\subsection{Dynamic Method}

As an extra work we also implemented another object detection model proposed by \cite{iccv}, and tested its performance on 8 categories. However due to time limitation, we could not perform a broad hyper-parameter study on this approach. We trained the dynamic model on 8 object classes of VOC2012 dataset, for 7 epochs (each epoch 20 steps) and tested the model (with maximum 10 steps for each image). The results can be seen in table \ref{class_d}. 

\begin{table}[htbp]
\caption{Different Object Categories}
\begin{center}
\begin{tabular}{|c|c|}
\hline
\textbf{Object class} & \textbf{ Average IoU}\\
\hline 
aeroplane &  0.3723 \\
\hline
bicycle &  0.1990\\
\hline
car &  0.1952\\
\hline
cat &  0.4115\\
\hline
bird &  0.2059\\
\hline
bus &  0.3858\\
\hline
dog &  0.2828\\
\hline
horse & 0.3375\\
\hline
\end{tabular}
\label{class_d}
\end{center}
\end{table}

\section{Conclusions}

In this project, we implemented two algorithms of object detection with deep reinforcement learning: a hierarchical model and a dynamic model. By adjusting hyperparameters and some setups, we achieved a hierarchical model with a reasonable performance in object detection. We also implemented a dynamic model, which has more freedom in its action setting (shape of bounding boxes) compared to hierarchical model. 

Comparing the two models, hierarchical method trains and performs faster than dynamic method, as it has fewer number of actions (6 vs. 9). It also fits the smaller objects better than dynamic method, since at each step the size of predicted box becomes smaller. However, making windows smaller at each time step is a weakness of this method, especially for bigger targets. Indeed, hierarchical model cannot recover from a bad region choice in the first step. It stays in the same region and choose its descendent sub-regions at each step. In Dynamic method on the other hand, the predicted box is able to move freely in image and is not bounded to any predefined region. The window is also able to become bigger at any step. One other weakness of hierarchical method is that the predicted box should always keep the shape of the original image, as all predefined sub-regions are a ratio of original image. In fact, if the image is vertical, the predicted boxes are all vertical, which may not fit the target object shape in image. This problem can be clearly seen in images included in the appendix.  

In terms of better accuracy, we are not able to make a firm statement that which model is more accurate, as there are some differences in the default settings of the two model and due to time limitation we could not build a fair comparison setting to compare the two methods. 

Both methods are dependent on the hyperparameters and variables chosen. As a result, to achieve good performance with these reinforcement learning object detection models, a careful design of hyperparameters should be considered. Besides, we are surprised to find that some of the techniques proposed by the papers, such as the action history vector included in the state representation, do not contribute to the performance of model in our case.

We also observed a peculiarity in hierarchical method training, that the best models are usually obtained in the first 5 to 7 epochs, and then the performance was decreased. This has conflict with the authors claim that they trained the model for 50 epochs. We suspect that our model is possibly under-trained and need to be trained on more data. In the original papers \cite{iccv} and \cite{Bellver2016HierarchicalOD}, the agent is trained on both VOC2012 and VOC2007 dataset; but due to limited computational resources, we only used VOC2012. Hence, we would like to combine the two datasets and retrain our models on them in future. 

The algorithms we implemented in this project are the first algorithms that attempt to treat the object detection problem as a dynamic decision process. The idea of formulating the process of locating the bounding box as a Markov decision process (MDP) is a breakthrough, which is quite different from the conventional neural network approaches applied in the computer vision field. As a result, even though the accuracy of our models cannot reach that of the current state-of-art algorithms on this task, it is still worthwhile to look into and understand these models to examine the potential of reinforcement learning in the future of computer vision.

Moreover, as we discussed in the method section, these models are currently class-dependent, which means that the agent can only search for one class of objects at a time. As a future direction, it would be meaningful to extend these models to be class-independent so that it can be trained with arbitrary images and find object of all classes at the same time. Furthermore, the current algorithms are not robust in finding multiple objects in image at the same time; this also can be addressed as a future work. 

\section{Links to code base and video}

We included the notebooks of our experiments, our trained models, and other contents of reproducibility checklist in Mega cloud storage, which is accessible through this link: 
\url{https://mega.nz/folder/7UMRmKAQ#PiPJ6RquyeWX48G3R0Eo-Q }

Our spotlight video is also available at:
\url{https://youtu.be/dcGP9mDnFf0 }

\section{Statement of Contribution}

Both Ruofeng and Manoosh worked on hyperparameter search, improving the models, interpreting the results, and writing the report. Manoosh also worked on coding and visualizing the results. Ruofeng also worked on preparing slides and spotlight video.

\bibliographystyle{plain}
\bibliography{references}

\section{Appendix}

Some good predictions by hierarchial model:

\begin{figure}[!htb]
  \centering
  \includegraphics[width=0.27 \textwidth]{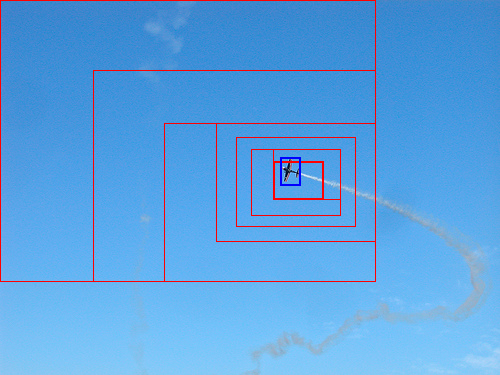}
  \includegraphics[width=0.27 \textwidth]{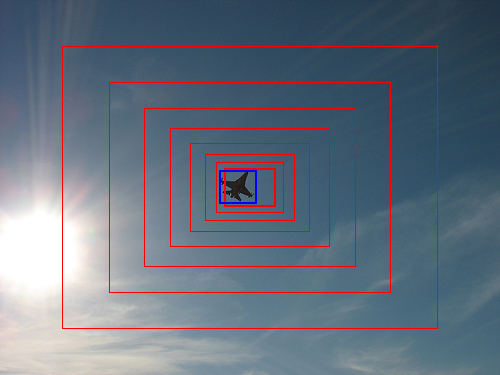}
  \includegraphics[width=0.27 \textwidth]{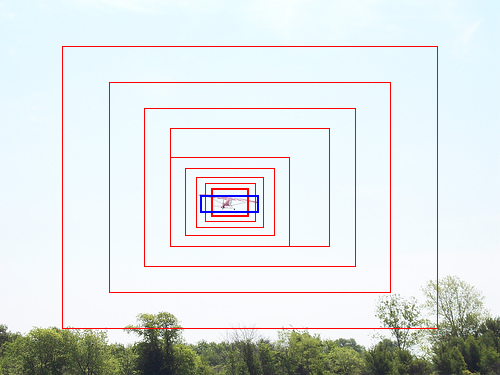}
  \includegraphics[width=0.27 \textwidth]{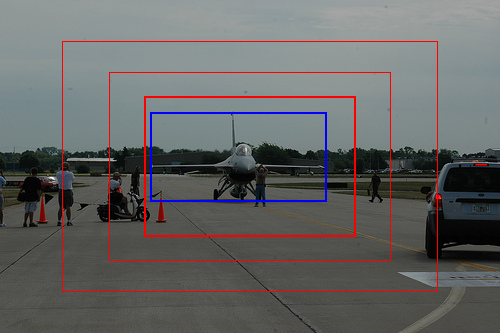}
  \includegraphics[width=0.27 \textwidth]{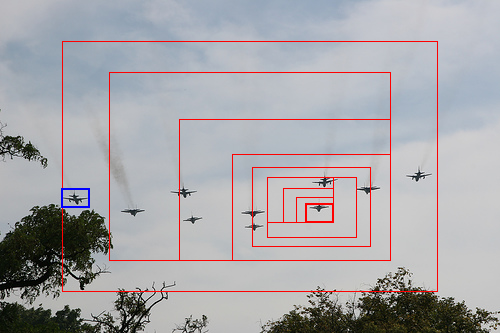}
  \includegraphics[width=0.24 \textwidth]{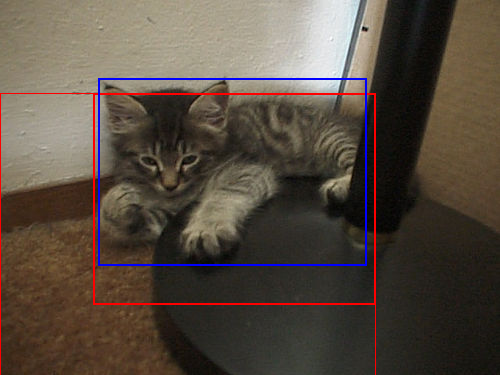}
  \includegraphics[width=0.24 \textwidth]{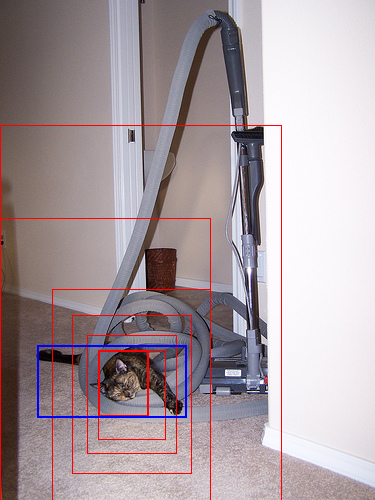}
  \includegraphics[width=0.25 \textwidth]{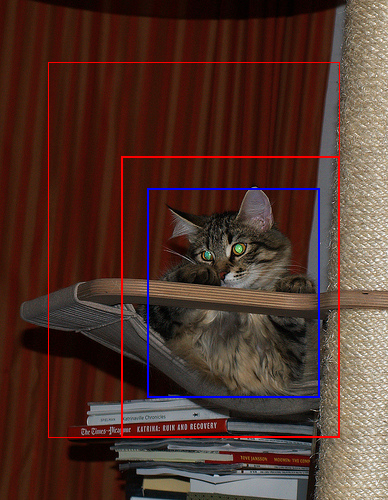}
  \includegraphics[width=0.27 \textwidth]{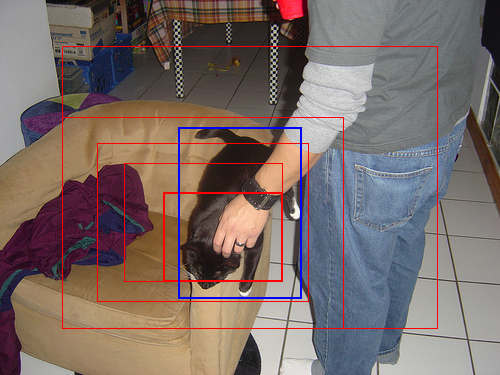}
  \includegraphics[width=0.27 \textwidth]{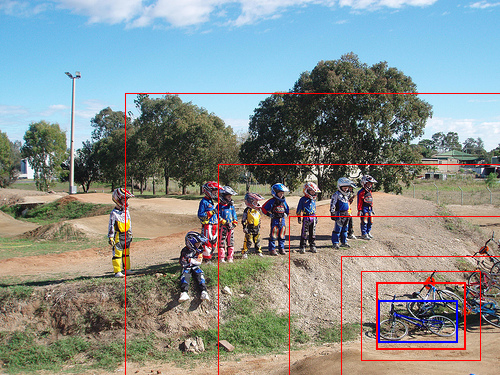}
  \includegraphics[width=0.27 \textwidth]{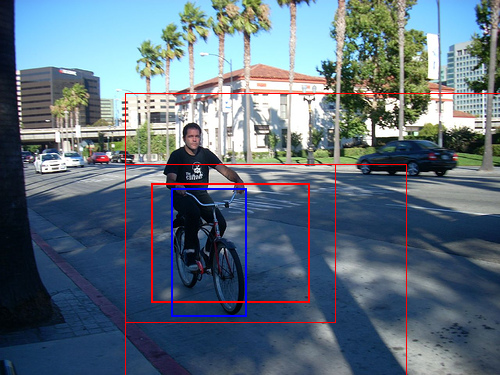} 
  \includegraphics[width=0.27 \textwidth]{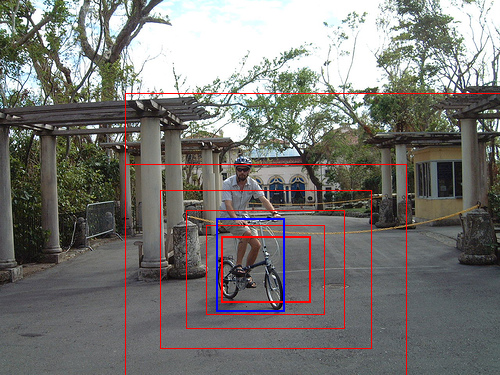}
  \caption{ Some good predictions by hierarchial model, all of the predictions are obtained by a model with fixed target object setting. The blue bounding box is a sample ground truth box for a target object, the red bounding boxes are the search path and the bold red bounding box is the final predicted box.}
  \label{fig:action}
\end{figure}

Some good predictions by dynamic model:

\begin{figure}[!htb]
  \centering

  \includegraphics[width=0.3 \textwidth]{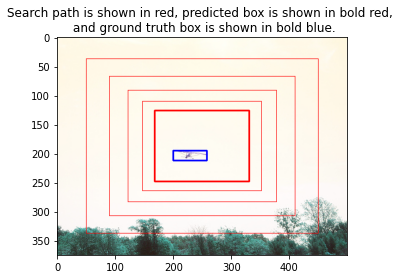}
  \includegraphics[width=0.3 \textwidth]{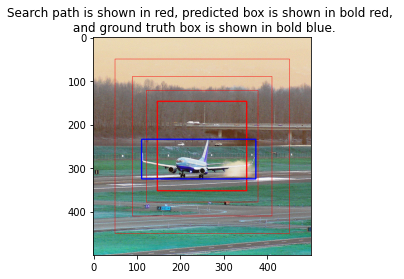}
  \includegraphics[width=0.3 \textwidth]{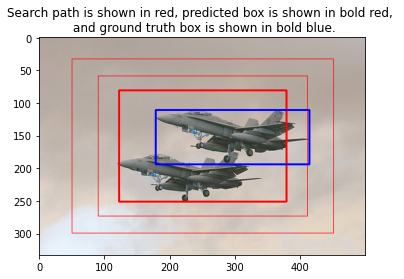}
  \includegraphics[width=0.3 \textwidth]{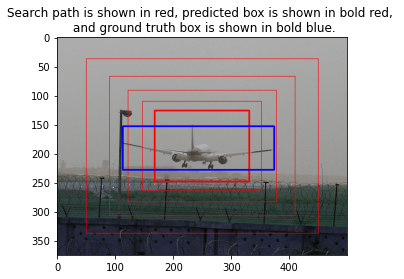}
  \includegraphics[width=0.3 \textwidth]{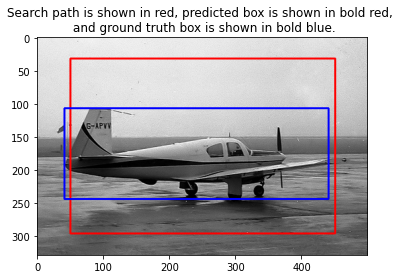}
  \includegraphics[width=0.3 \textwidth]{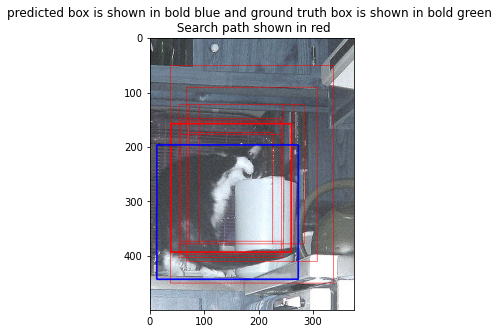}
  \includegraphics[width=0.3 \textwidth]{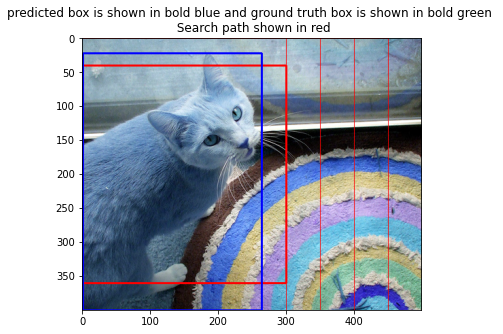}
  \includegraphics[width=0.3 \textwidth]{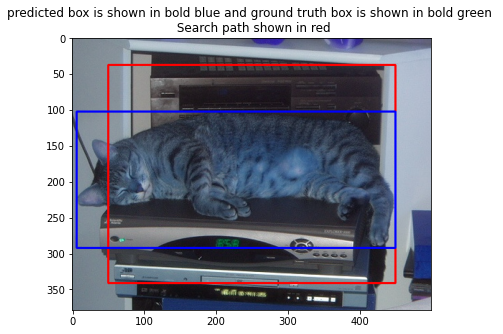}
  \includegraphics[width=0.3 \textwidth]{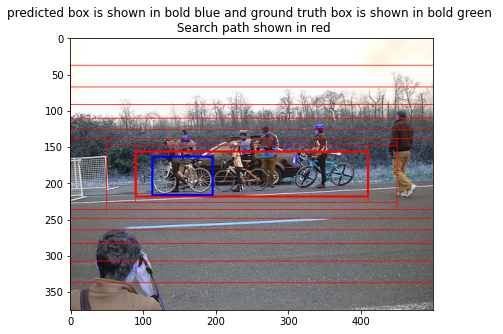}
 \includegraphics[width=0.3 \textwidth]{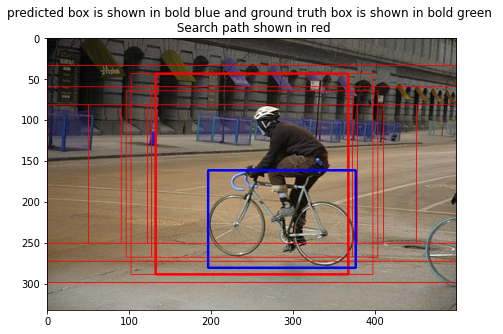}
  \includegraphics[width=0.3 \textwidth]{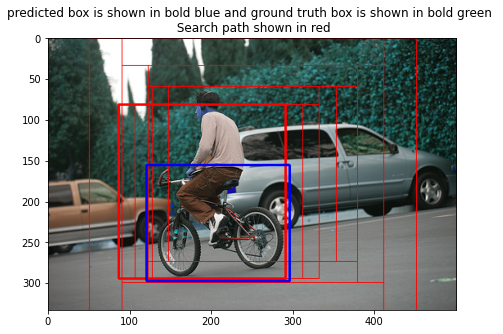}
  \includegraphics[width=0.3 \textwidth]{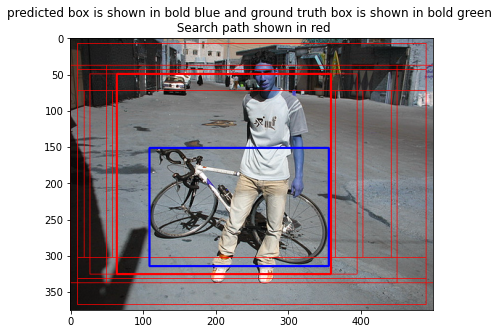}
  \caption{ Some good predictions by dynamic model. The blue bounding box is a sample ground truth box for a target object, the red bounding boxes are the search path and the bold red bounding box is the final predicted box.}
  \label{fig:action}
\end{figure}

\end{document}